\documentclass[letterpaper, 10 pt, conference]{ieeeconf}
\IEEEoverridecommandlockouts                               
\usepackage{cite}
\usepackage[utf8]{inputenc}
\usepackage{amsmath}
\usepackage{amsfonts}
\usepackage{amssymb}
\usepackage{mathtools}
\usepackage{algorithm,algorithmic}
\usepackage{graphicx}
\usepackage{textcomp}
\usepackage{xcolor}
\usepackage{hyperref}
\usepackage[export]{adjustbox}
\usepackage[lofdepth,lotdepth]{subfig}
\usepackage{footmisc}

\def\BibTeX{{\rm B\kern-.05em{\sc i\kern-.025em b}\kern-.08em
    T\kern-.1667em\lower.7ex\hbox{E}\kern-.125emX}}

\begin{document}
    
\pubid{\makebox[\columnwidth]{\copyright 2019 IEEE \hfill} \hspace{\columnsep}\makebox[\columnwidth]{ }}

\title{\LARGE \bf Contract Statements Knowledge Service for Chatbots}

%


\author{Boris Ruf$^{1}$,
Matteo Sammarco$^{1}$,
Marcin Detyniecki$^{1,2,3}$
\thanks{$^{1}$Boris Ruf and Matteo Sammarco are with AXA, Paris, France {\tt\small \{boris.ruf,matteo.sammarco\}@axa.com}}
\thanks{$^{2}$Marcin Detyniecki is with AXA, Paris, France and with Sorbonne Universit\'e, CNRS, LIP6, Paris, France and with Polish Academy of Science, IBS PAN, Warsaw, Poland {\tt\small marcin.detyniecki@axa.com}}}

\maketitle
\begin{abstract}
Towards conversational agents that are capable of handling more complex questions on contractual conditions, formalizing contract statements in a machine readable way is crucial. However, constructing a formal model which captures the full scope of a contract proves difficult due to the overall complexity its set of rules represent. Instead, this paper presents a top-down approach to the problem. After identifying the most relevant contract statements, we model their underlying rules in a novel knowledge engineering method. A user-friendly tool we developed for this purpose allows to do so easily and at scale. Then, we expose the statements as service so they can get smoothly integrated in any chatbot framework. 
\end{abstract}


\section{Introduction}
For a long time, researchers in artificial intelligence (AI) have been intrigued by the idea of developing a conversational agent that is capable of having a coherent conversation with humans~\cite{Weizenbaum1966,Colby1971,Wallace2009}. Recent breakthroughs in semantics and speech recognition have given rise to hopes for robust solutions to the problem~\cite{Mesnil2015,Amodei2015}. Major information technology companies have released digital assistants and chatbot frameworks to facilitate the building of conversational agents~\cite{Lopez2018,Janarthanam:2017:HCC:3203597}.

However, research studies on the user perception and expectations from users of chatbots indicate that the systems still require significant improvements in order to provide a meaningful experience~\cite{Zamora2017}. Also, demand analysis identified a need for specialized digital assistants in customer facing processes, in particular in the insurance sector~\cite{DBLP:journals/corr/abs-1812-07339}. A promising approach to advance in this field are comprehensive knowledge engineering methodologies which back the chatbot and upgrade its conversational capabilities from small talk to domain expert~\cite{Cameron2018}.

\begin{figure}[t]
\begin{center}
    \includegraphics[width=1\linewidth]{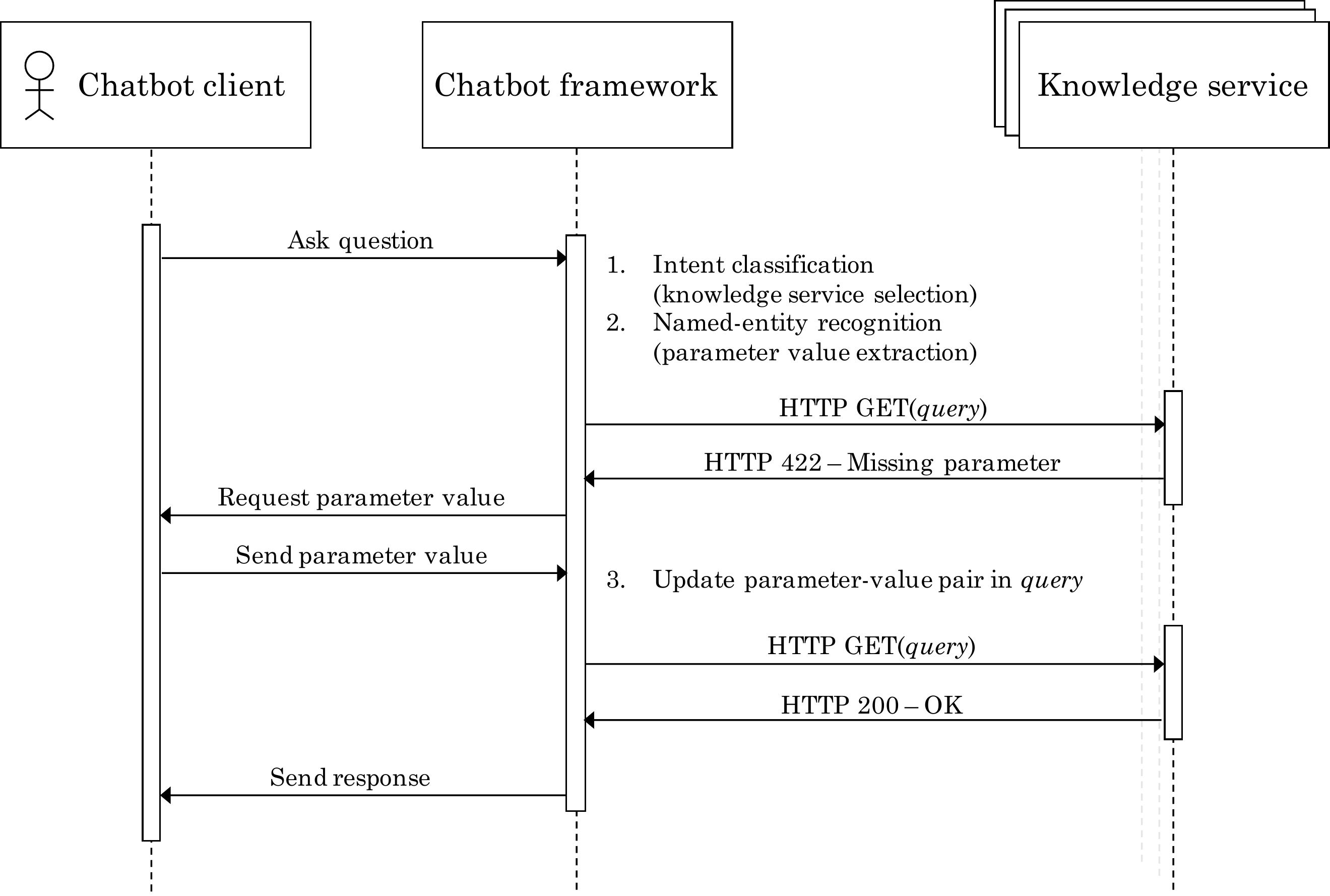}
    \caption{A sequence diagram of a sample conversation flow. The knowledge service incorporates the contract statement.}
    \label{fig:sequence}
\end{center}
\end{figure} 

In our case, we are interested in the domain of contract statements of insurance policies. In order to narrow down the problem, we focus on the kind of questions that can be answered based on linear rules which take a finite set of parameter-value pairs as input. We assume that most customer support inquiries fall in the domain of our interest, i.e. questions around coverage and pricing in real-world insurance policies.

Modeling contracts bottom-up as self-executing contracts, also known as ``smart contracts", is appealing due to its large potential of automation and its resilience to tampering~\cite{Bartoletti2017}. However, it is extremely hard to apply this concept on real-world contracts based on existing contract law provisions because smart contracts are ``indifferent to the fundamental legal principles, such as lawfulness, fairness, and protection of the weaker party"~\cite{Savelyev2017}. 

Matching inquiries around a contract using a conventional FAQ bot would be an option, in particular when using a deep learning neural model for answer selection~\cite{Huang2018}. However, the use of static answers can quickly become complex and hard to maintain due to the arbitrary number of combinations. 

In this paper, we propose a knowledge engineering methodology to model contract statements which support the automated handling of such questions. We choose a top-down approach, manually selecting the most relevant statements of a contract and modeling their underlying rules in a straightforward and accessible manner. The rules are based on combinations of different parameter spaces. This allows for the modelling of complex statements while facilitating easy maintainability. Conversational agents can query the rules via API which allows for seamless integration with any chatbot framework. We also assess the theoretical statement expressivity of our approach and demonstrate that it holds a significant potential of complexity reduction compared with conventional FAQ bots.

The remainder of this paper proceeds as follows. We begin with an overview of the overall architecture we propose and introduce a toy scenario. Next, in Section~\ref{sec:knowledge_service}, we define the expected input to the system, describe the theory behind our concept of contract statement modeling, and outline the different possible query scenarios.
Then, Section~\ref{sec:complexity} studies the theoretical complexity reduction of our approach with respect to the maximum and the actual number of questions covered. Further, we provide the technical specification of our prototype which realises the present concept in Section~\ref{sec:implementation}.
Finally, we conclude by providing insights in possible future fields of research.
\begin{figure}[t]
\begin{center}
    \includegraphics[height=6cm]{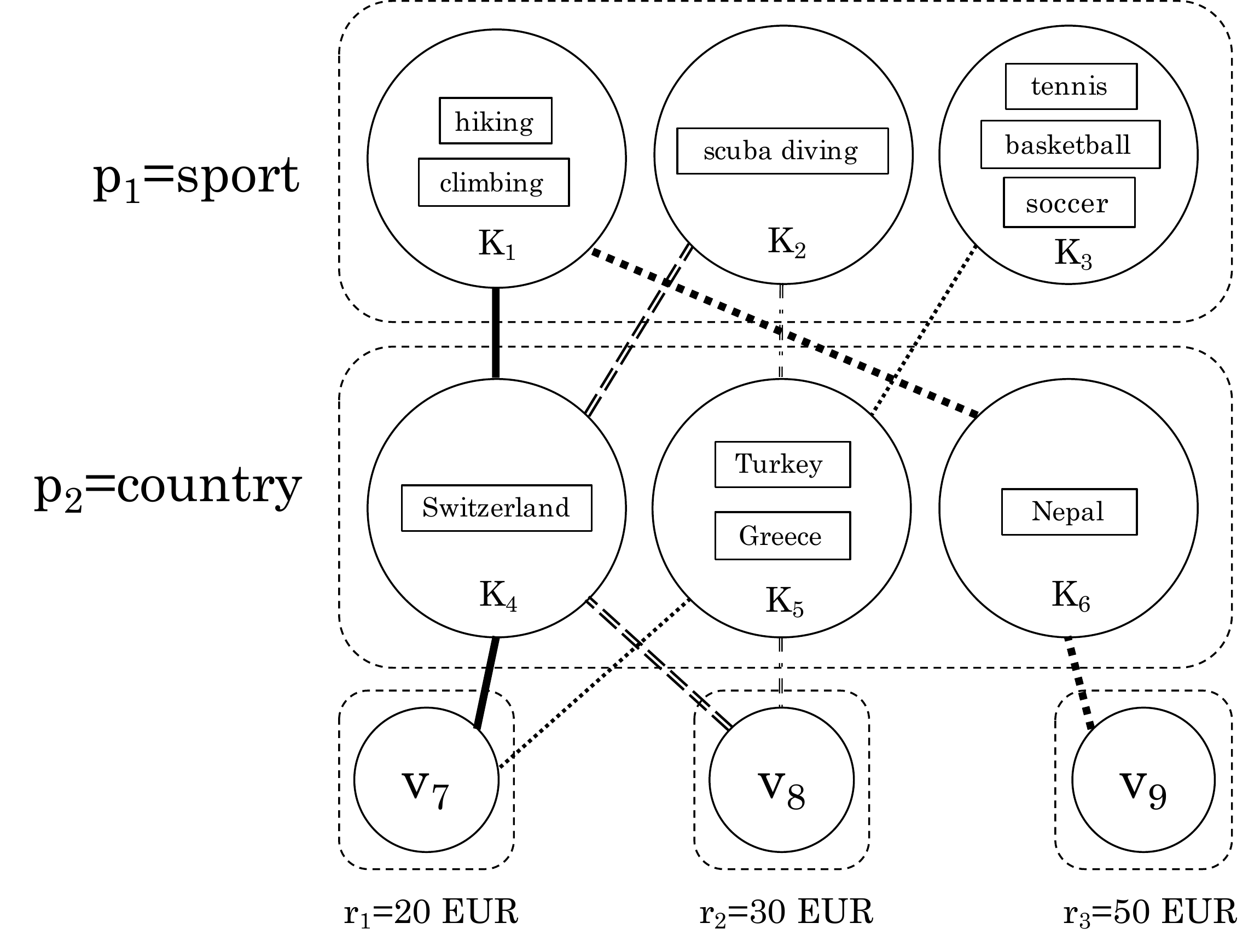}
    \caption{Toy scenario: The hypergraph represents a contract statement of a fictional accident insurance statement.}
    \label{fig:example}
\end{center}
\end{figure}

\section{General architecture}
\label{sec:general_architecture}

The proposed architecture to respond to customer support inquiries consists of a chatbot client, a chatbot framework and several knowledge services. The conversation flow between those components has been illustrated in a sequence diagram in Figure~\ref{fig:sequence}. 

\subsection{Chatbot client}
The chatbot client is the messaging interface provided to interact with the customer support. It allows to send and receive short messages and preserves the conversation in a local message log.

\subsection{Chatbot framework}
\label{sec:chatbot_framework}
The chatbot framework manages the conversation flow and preprocesses the inquiries. In this paper, we focus on designing the knowledge service, the operations of the chatbot framework are not in the scope of this paper. Thus, we only briefly describe them as follows. Every inquiry sent to the chatbot framework is preprocessed using Natural Language Processing (NLP) methods. First, a text classifier is used for intent classification. By understanding the matter of the inquiry, the framework can identify the corresponding knowledge service. Then, named-entity recognition is run on the user message to retrieve the parameter-value pairs. Finally, the parameter-value pairs are dispatched to the knowledge service as query parameters.

\subsection{Knowledge service}
Every knowledge service represents an endpoint for one specific type of question. It contains expert knowledge formalized as rules of a contract statement. The properties of the knowledge service are described in full length in Section~\ref{sec:knowledge_service}.

\subsection{Toy scenario}
\label{sec:toy_scenario}
Throughout this paper we will illustrate all aspects of our proposal with a simple toy scenario. Its model is shown in Figure~\ref{fig:example}. The subject is a fictional accident insurance which covers sports accidents in different geographical regions. Prices vary depending on the kind of sport and the country. We would like to automate responses to price requests based on the customer's parameters.

\begin{figure}[t]
\begin{center}
    \includegraphics[height=6cm]{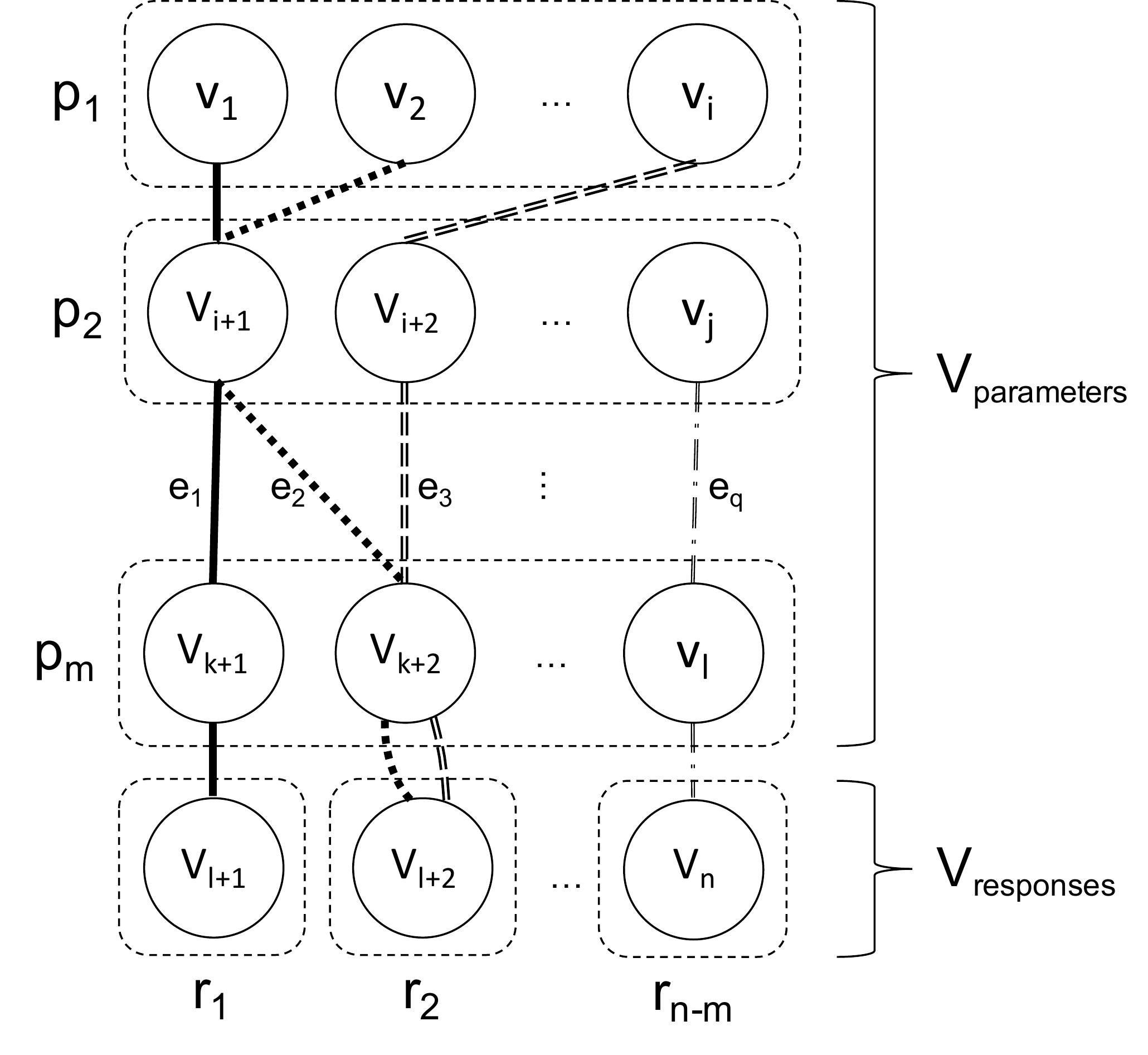}
    \caption{The schematic representation of a hypergraph $H$ which models a contract statement. The vertices $v_i$ represent the different sets of parameter values and also the possible responses. Rules are defined by hyperedges $e_i$ which connect several vertices.}
    \label{fig:schema}
\end{center}
\end{figure}

\section{Knowledge service}
\label{sec:knowledge_service}
The knowledge service plays the role of the domain expert. It processes the query, identifies the matching rule and delivers the appropriate answer. In the following, we describe the expected input to the service, the modeling of a contract statement, and the different query scenarios.

\subsection{Input query}
\label{sec:input}
The knowledge service expects a structured request containing a set of query parameters as input. As previously described, this request is prepared by the chatbot framework. 

In the context of our toy scenario, the customer could for example enter the question ``How much is an accident insurance for scuba diving in Turkey?" in the chatbot client. The chatbot framework classifies the intent of the message and match the corresponding knowledge service which handles pricing questions. Further, it identifies the required parameters for this service, here \textit{sport} and \textit{country}. Using named-entity recognition, the relevant parameter values, in this example ``scuba diving" and ``Turkey", are extracted and dispatched as query parameters to the knowledge service.

\subsection{Contract statement}
\label{sec:contract_statement}

We first propose to select relevant statements from a contract, that is, specific rules which are frequently subject of customer inquiries. We suggest to model such statements as hypergraphs. The vertices of a hypergraph represent the different parameter values and also the possible responses of each statement. Rules are defined by edges which connect several vertices.

In our toy scenario, the statement is about pricing. The vertices represent the possible parameter values, here groups of sports and countries. They also stand for the responses, in this example the different prices. The rules define the prices for each available combination of parameter values.

We choose a hypergraph to model a statement because, unlike a simple graph, a hypergraph has edges which can join any number of vertices. Let $H$ = $(V,E)$ be a finite, undirected hypergraph with vertices $V=\{v_1, v_2, ..., v_n\}$. $E$ is a set of hyperedges $\{e_1, e_2, ..., e_q\}$, such that each hyperedge contains an arbitrary set of vertices, $e_i \subseteq V$. More precisely, $E$ is a subset of $\mathcal{P}(V)$, where $\mathcal{P}(V)$ is the power set of $V$.

An schematic presentation of a hypergraph $H$ as defined above can be found in Figure~\ref{fig:schema}. An instance of this hypergraph representing the statement of the toy scenario is shown in Figure~\ref{fig:example}.

\subsubsection{Vertices}
The vertices are split in 2 different types: the ones that represent the parameter values, and the ones that stand for the available responses. 

Let vertices $\{v_1, v_2, ..., v_l\}, 1 \leq l \leq n$ be the first subset $V_{parameters}$ of $V$. They are each labeled with a string denoted as parameters $P=\{p_1, ..., p_m\}$ where $p_i \in S$. $S$ is the set of strings which consist of at least 2 characters. In our toy scenario, the parameters are \textit{sport} and \textit{country}. 

The labels are unique, $p_i \neq p_j$, and the maximum number of parameters is the total number of vertices, thus $1 \leq m \leq n$. Each vertex of $V_{parameters}$ represents a set of possible values for the given parameter. We describe those values as set of keywords $K_i=\{k_1, k_2, ..., k_{n_{i}}\}$ which consist of at least one string, thus where $k_j \in S, n_i \in \mathbb{N}$. The keywords are required to be unique per parameter, thus $k_x \neq k_y$ for each $v_i \in V_{parameters}$ where $label(v_i)=label(v_k).$ Otherwise, several rules could apply simultaneously, and there would be no unambiguous result. In our continuous example, the keywords are $K_1=\{``hiking", ``climbing"\}$, $K_2=\{``scuba~diving"\}$, etc.

Let vertices $\{v_{l+1}, v_{l+2}, ..., v_n\}$ be the second subset $V_{responses}$ of $V$, labeled with responses $R=\{r_{1}, r_{2},..., r_{n-m}\}$ where $r_i \in S$. Responses are to be unique in our model, thus $r_i \neq r_j$. In the toy scenario, the responses are $R=\{``20~EUR", ``30~EUR", ``50~EUR"\}$.

\subsubsection{Edges}
Edges represent the statement's rules. We require the rules to take at most one value per parameter, which is why we only allow edges to connect vertices of different parameters. Formally, this means that each hyperedge $e_i \in E$ must contain vertices $v_x$ and $v_y$ so that $label(v_x)$ $\neq$ $label(v_y)$. Also, it must include a minimum of 2 vertices, including 1 vertex of $V_{responses}$, so that $2\leq|e_i|\leq|P|+1$. 


For example, in our toy scenario displayed in Figure~\ref{fig:example}, the rules show that the prime for $``hiking"$ in $``Switzerland"$ would be cheaper than for $``hiking"$ in $``Nepal"$. On the other hand, insurance for $``scuba~diving"$ in $``Nepal"$ is not offered. 


\begin{algorithm}[t]
\begin{algorithmic}
   \STATE{\bfseries Input:} 
     $H(V_{parameters}+V_{responses}, E)$\\
     $x \in \{(p_1, k_1), (p_2, k_2), ..., (p_n, k_n)\}$ where $p_i \in P, k_i \in S$ 
   \STATE{\bfseries Output:} $response, http\_code$
   \IF{$k_i$ valid}
     \STATE $V_{matches} \leftarrow$ Identify all vertices $v_i \in V_{parameters}$ which are labeled with $p_i$ and include $k_i$ in the associated set of keywords
     \STATE $edge \leftarrow$ Identify edge $e_i \subseteq V_{matches}, e_i \in E$
     \IF{$edge$ exists} 
       \STATE{$v_{r} := edge \subseteq V_{responses}$} 
       \STATE{$response := label(v_{r})$} 
       \STATE{$http\_code := 200$}
     \ELSE
       \IF{$n < |P|$} 
         \STATE{$response := p_{n+1}$}
         \STATE{$http\_code := 422$}
       \ELSE
         \STATE {$response := false$}
         \STATE{$http\_code := 200$}
       \ENDIF
     \ENDIF
   \ELSE
     \STATE {$response := false$}
     \STATE{$http\_code := 400$}
   \ENDIF
\end{algorithmic}
\caption{Outline rule matching algorithm}
\label{algorithm}
\end{algorithm}

\subsection{Query scenarios}
\label{sec:rule_matching}
The necessary steps to find the right response to an inquiry have been outlined in the form of pseudo code in Algorithm~\ref{algorithm}. First, we identify all vertices of $V_{parameters}$ which match the parameter-value pairs provided in the query. Then, we search for a hyperedge that connects those vertices. For this purpose, we consider the following scenarios.

\subsubsection{Complete input query (default)}

If parameter values were provided for all parameters, and a fully connected hyperedge exists, we single out the vertex which belongs to the subset $V_{responses}$. Notice that by definition each hyperedge must include exactly one such vertex. As response we send the label of this vertex to the chatbot framework.

If the matched vertices are not fully connected by a hyperedge, it means that none of the defined rules in the statement apply. Thus, the response is the boolean value $false$. Depending on the use case, this response may be interpreted as ``no", or as ``undefined" – impossible to make a valid statement. 

\subsubsection{Incomplete input query}
If the query is incomplete, a fully connected hyperedge may still exist, since we do not require hyperedges to include vertices of every single parameter. In this case, we send the label of the vertex which belongs to the subset $V_{responses}$, just as described above. 

However, if the matched vertices are not represented by a fully connected hyperedge, the knowledge service prompts the user for the missing parameter values until either a hyperedge is found or the query is complete and contains values for all parameters. More specifically, the label of the first parameter $p_i \in P$ without value $k_i$ is sent to the chatbot framework with HTTP status code 422 (``Unprocessable Entity"). The framework requests the parameter value from the user, and finally resubmits the updated parameter query. This approach enhances the knowledge service with a very basic capacity to manage part of the conversational flow.

For example, if the question ``How much is an accident insurance for scuba diving?" was asked in the toy scenario, the knowledge service would respond with ``country" and HTTP code 422, prior to providing a final response. The chatbot framework could rephrase the request to ``In which country would you like to exercise the sport?", include the answer of the user to the query and pass the completed parameter-value pair to the knowledge service.

\subsubsection{Invalid input query}
A parameter query may include unexpected input, i.e. multiple values for the same parameter. For the sake of simplicity, the knowledge service currently only detects such exceptions and leaves it to the chatbot framework to handle them appropriately. More concretely, the knowledge service responds with the boolean value $false$ and the HTTP status code 400 (``Bad Request").
\begin{figure}[t]
\begin{center}
    \includegraphics[width=1\linewidth]{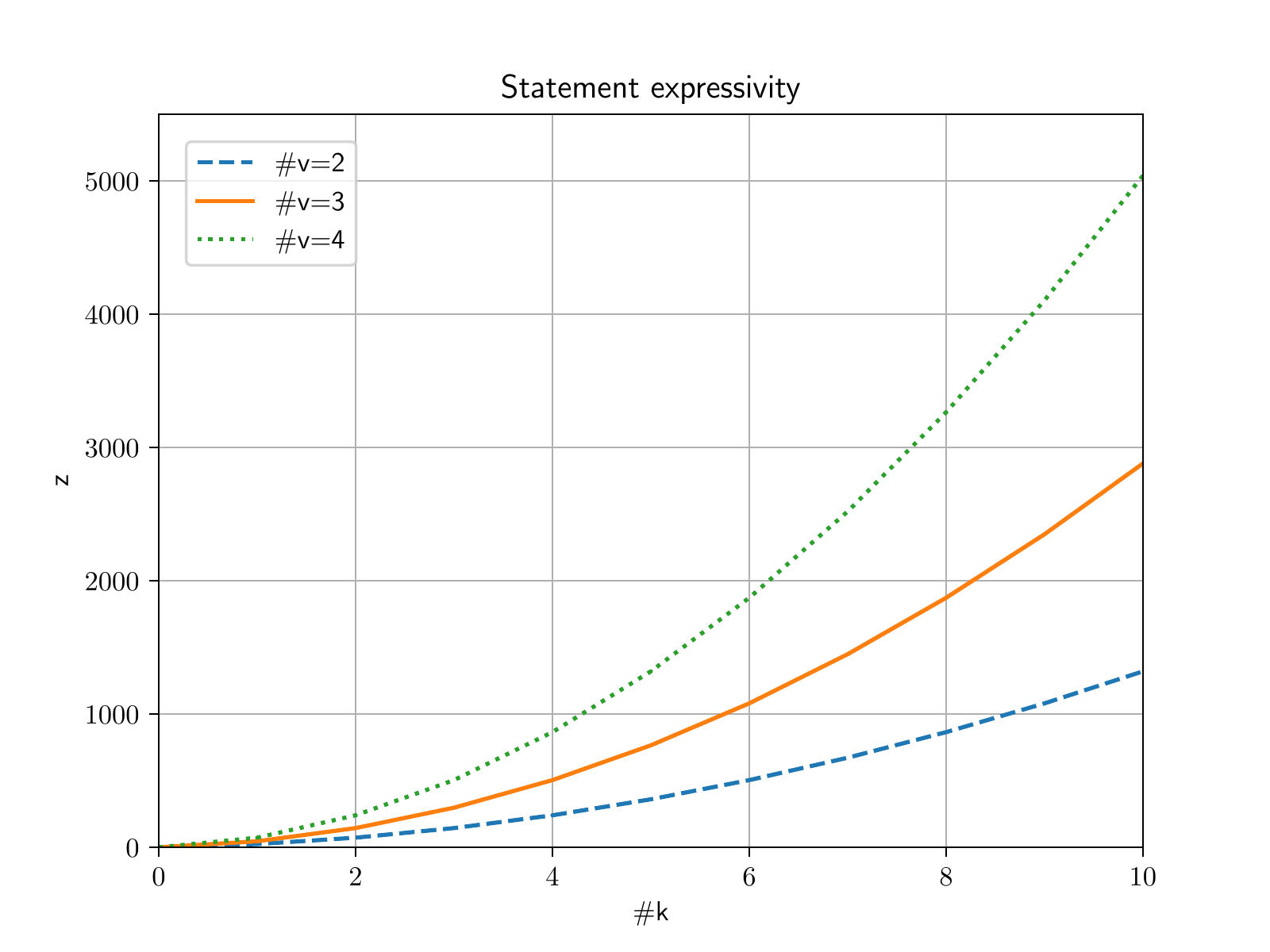}
    \caption{Statement expressivity scenarios for a differing number of vertices per parameter (\#v) over an increasing number of keywords per vertex (\#k). The number of parameters $|P|$ and the number of response vertices $|R|$ are fixed to 2 and 3 respectively for better comparability.}
    \label{fig:statement_expressivity}
\end{center}
\end{figure}

\begin{figure*}
  \begin{center}
    \subfloat[Interactive hypergraph: The vertices are displayed as boxes. The 2 top layers contain the ones which represent parameter values, the bottom layer contains the response vertices. Edges of different color constitute the different rules.]{
      \includegraphics[frame,width=0.46\textwidth]{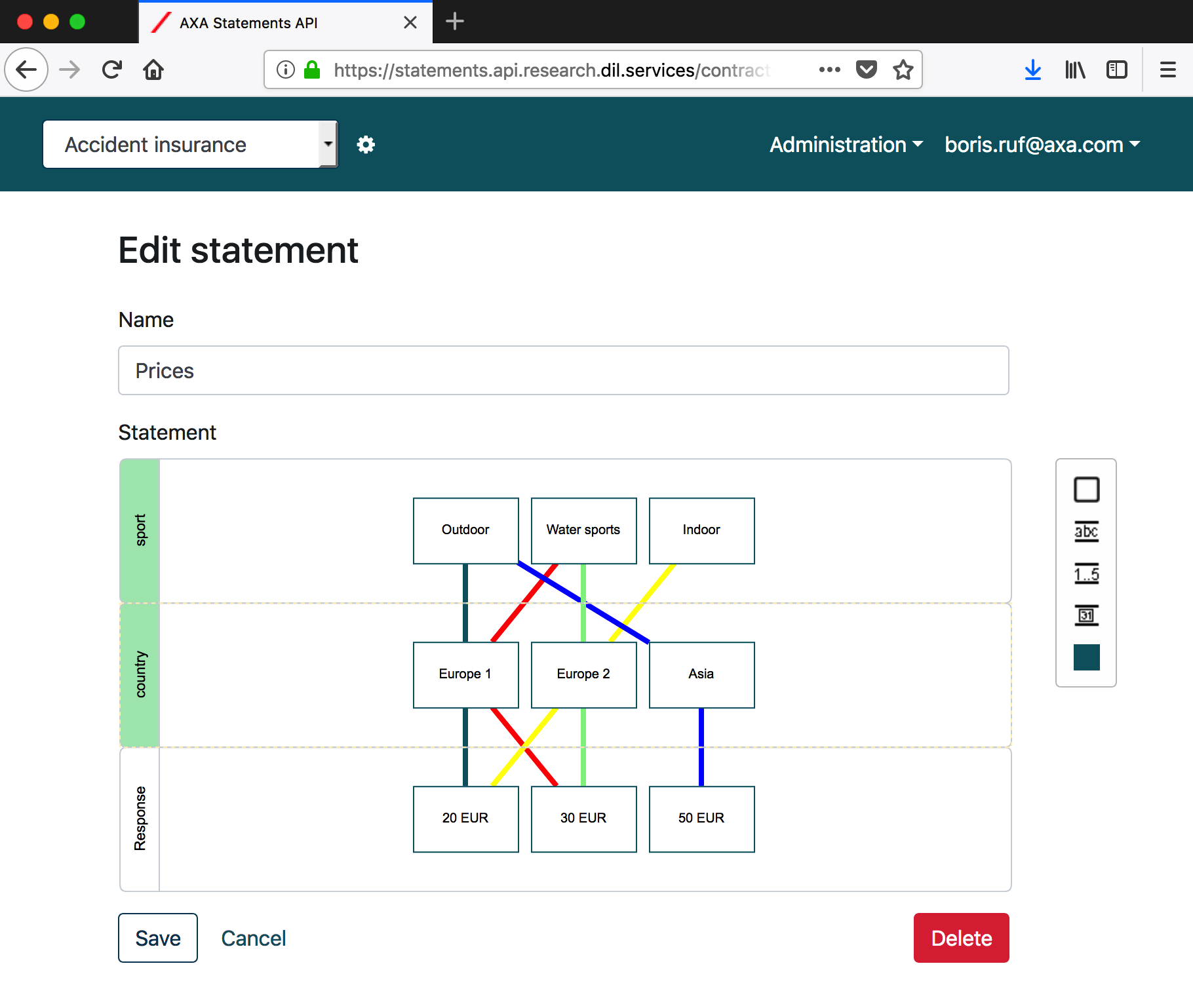}
      \label{sub:screenshot1}
    }
    \hspace{0.2cm}
    \subfloat[Keywords: For each vertex which represents a set of parameter values, matching keywords can be entered in the format of tags.]{
      \includegraphics[frame,width=0.46\textwidth]{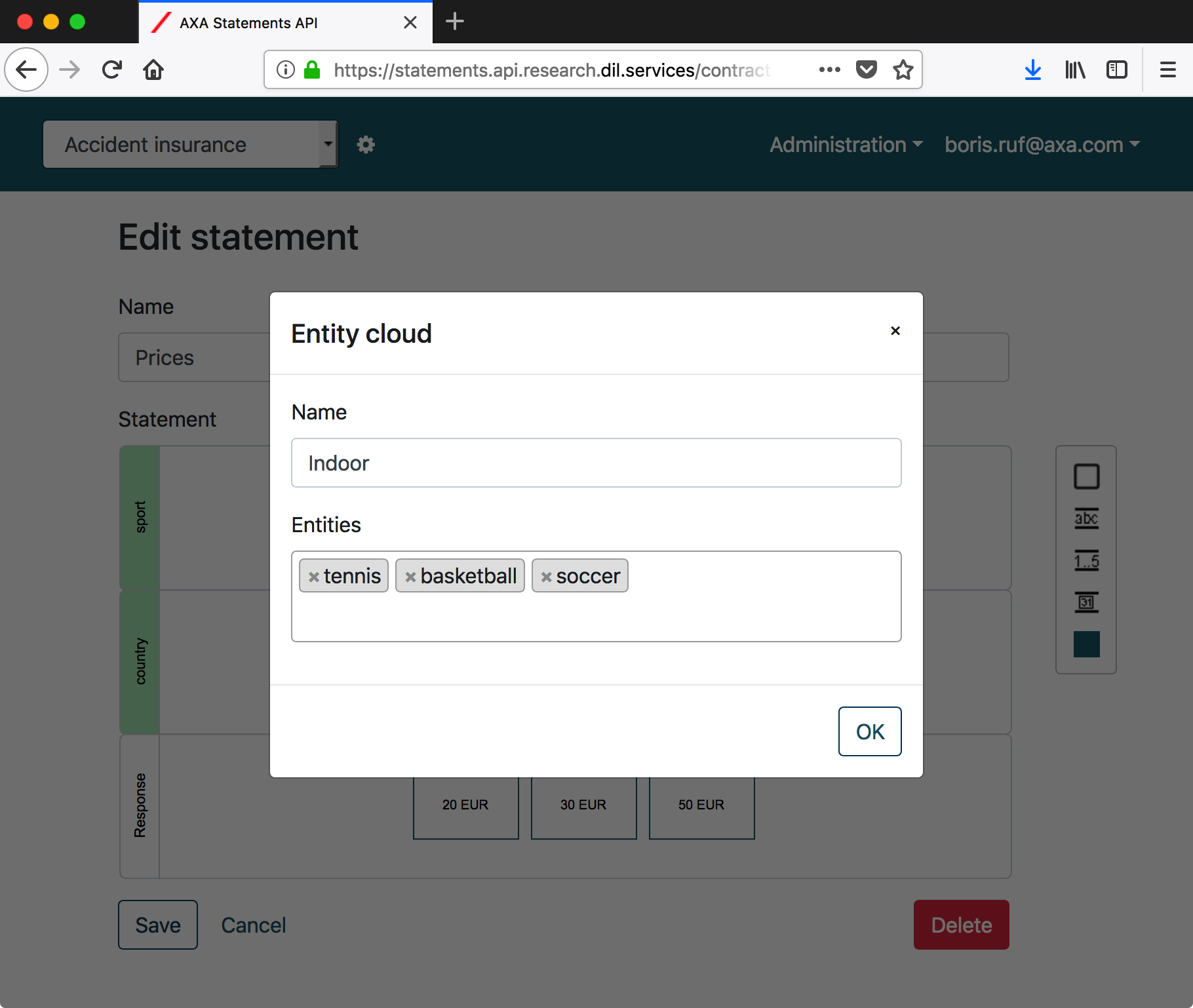}
      \label{sub:screenshot2}
    }
    \caption{Graphical user interface of the prototype}
    \label{fig:screenshots}
  \end{center}
\end{figure*}

\section{Complexity reduction}
\label{sec:complexity}

In theory, the previously presented type of questions could be answered by conventional FAQ bots, too. However, such frameworks would require to write one explicit answer for every single question. This would result in a very large set of question-answer pairs, which is time-consuming to achieve and difficult to maintain. On the contrary, the contract model presented in this paper induces that each statement represented by a hypergraph factorizes all possible answers into rules and as such serves as dynamic answer to multiple questions.

\subsection{Statement expressivity}
To estimate the maximum number of different questions that could be covered by a hypergraph, we need to add up all possible parameter-value pair combinations for any possible hyperedge. To achieve this, we first count the keywords represented by each vertex of $V_{parameters}$ labeled with the same parameter $p_i \in P$ and assign those numbers to a set $\sigma$. Then, we can compute with the following formula the value of $z$ which represents the maximum number of conventional, static question-answer pairs the present statement replaces:

\begin{equation}
z = \sum_{s \in \mathcal{P}(\sigma)}\prod_{c \in s}c
\label{eq:statement_expressivity}
\end{equation}

where $\mathcal{P}(\sigma)$ is the power set of $\sigma$, and $c$ is the element of each subset of $\sigma$. 

We observe that the size of $|K_i|$, which corresponds to the number of values accepted for one parameter vertex $v_i$, as well as the total number of parameter layers $|P|$, have a strong influence on the overall complexity.

For the toy scenario, the value of $z$ is calculated as follows:
\begin{eqnarray*}
\begin{aligned}
\sigma =&{} \{(|K_1|+|K_2|+|K_3|),(|K_4|+|K_5|+|K_6|)\}\\
            =&\{(2+1+3),(1+2+1)\}\\
            =&\{6,4\}\\
\mathcal{P}(\sigma)  =& \{\{6\},\{4\},\{6,4\}\}\\
z =& 6+4+6\cdot4\\
 =& 34
\end{aligned}
\end{eqnarray*}

As result, up to 34 conventional question-answer pairs can be mapped to this single hypergraph with 3 levels and 9 nodes.

\subsection{Total questions covered by a specific statement}
In order to calculate the number of standard FAQ bot questions covered by a single given hypergraph, we sum up all possible parameter-value pair combinations for each hyperedge present in the graph. To achieve this, we first identify all vertices $v_i \in V_{parameters}$ included in a hyperedge. Then, we count the keywords represented by each of those vertices and multiply the results. In order to compute $t$, the total number of questions covered by the hypergraph, we repeat this procedure for all remaining hyperedges, and add up the sums. The resulting formula is:

\begin{equation}
t = \sum_{e \in E}\prod_{v= e \subseteq V_p}|K_{index(v)}|
\label{eq:total_questions}
\end{equation}

where $E$ is the set of hyperedges, and $V_p=V_{parameters}$.

For example, in the case of the toy scenario illustrated in Figure~\ref{fig:example}, the value of $t$ is computed as follows:

\begin{eqnarray*}
\begin{aligned}
t =&{} |K_1|\cdot|K_4|+|K_1|\cdot|K_6|+|K_2|\cdot|K_4|\\
  & +|K_2|\cdot|K_5|+|K_3|\cdot|K_5|\\
  =& 2\cdot1+2\cdot1+1\cdot1+1\cdot2+3\cdot2\\
  =& 13
\end{aligned}
\end{eqnarray*}

Hence, the statement modelled in the toy scenario substitutes 13 conventional FAQ bot questions.


\subsection{Impact of large number of keywords}
It is worth to note that with $z$ we only estimate the upper bound, and $t$, the number of actual total questions covered, may be considerably lower. However, in real-world scenarios the number of keywords (\#k) can easily reach much higher values than illustrated in the toy scenario, for example when the keywords represent enumerations of cities or countries. 

In Figure~\ref{fig:statement_expressivity}, we display varying statement expressivity scenarios. Concretely, we plot the maximum number of questions that could be covered for 3 different configurations. This value, $z$, is computed as defined in Equation~\ref{eq:statement_expressivity}, with respect to the number of keywords per vertex (\#k). In order to investigate the impact of one variable on $z$, we keep the number of parameters $|P|$ fixed to 2, and the number of response vertices $|R|$ fixed to 3. In the 3 different scenarios, we set the number of vertices per parameter (\#v) to 2, 3 and 4.

In general, we observe that the potential number of questions which can be covered by one statement grows exponentially over the number of keywords per vertex (\#k) at high rates. Even for the very low values of \#v, $|P|$ and $|R|$ chosen for the scenarios, $z$ rapidly reaches the four-digit range. Further, we see that a higher number of vertices per parameter (\#v) increases the growth rate significantly. 
\section{Implementation}
\label{sec:implementation}

Our prototype\footnote{Source code available at\\ \url{https://github.com/axa-rev/reasoning-api-framework}} is a multi-tenant web application implemented in Ruby on Rails, a web framework built on top of the object-oriented scripting language Ruby~\cite{RubyOnRails}. 

The graphical user interface is backed by the Bootstrap UI framework, and enhanced with the JavaScript diagramming library mxGraph~\cite{Bootstrap,mxGraph}. Figure~\ref{sub:screenshot1} shows the interactive user interface which is used to design a contract statement in the form of a hypergraph with vertices and edges. The vertex labels have been implemented as stacked layers. The 2 top layers contain the parameter values, the layer on the bottom the response vertices. Hyperedges can be created by simply drawing edges between the vertices. Different edge colors represent different hyperedges. The set of keywords associated with a vertex can be easily updated as displayed in Figure~\ref{sub:screenshot2}.

\begin{figure}[t]
  \begin{center}
    \subfloat[Correct]{
      \includegraphics[width=0.2\textwidth]{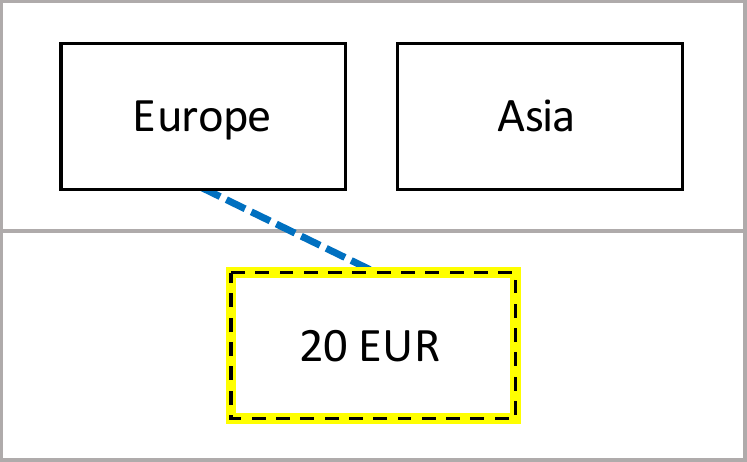}
      \label{sub:validation1}
    }
    \vspace{0.5cm}
    \subfloat[Wrong: Self-reference]{
      \includegraphics[width=0.2\textwidth]{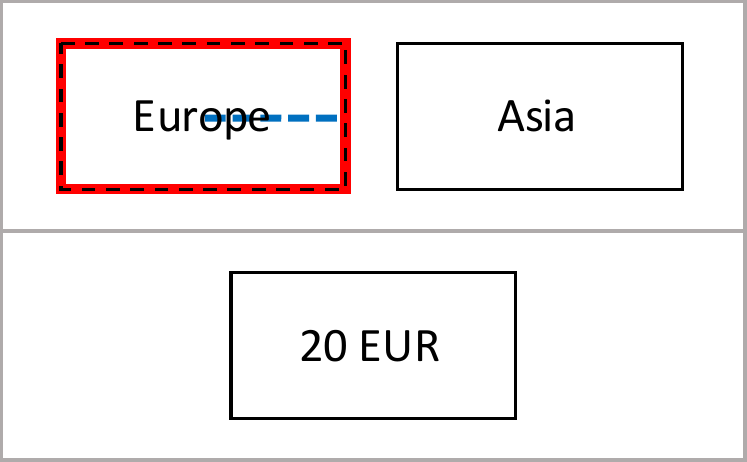}
      \label{sub:validation2}
    }
    \vspace{0.5cm}
    \subfloat[Wrong: Connection within same layer]{
      \includegraphics[width=0.2\textwidth]{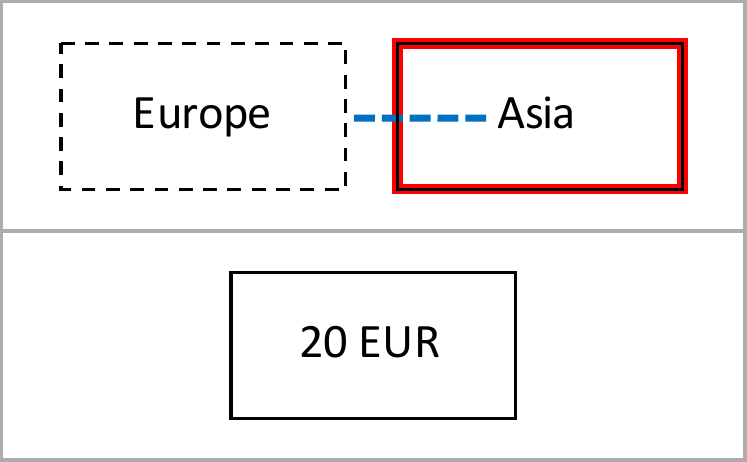}
      \label{sub:validation3}
    }
    \vspace{0.5cm}
    \subfloat[Wrong: Rule already exists]{
      \includegraphics[width=0.2\textwidth]{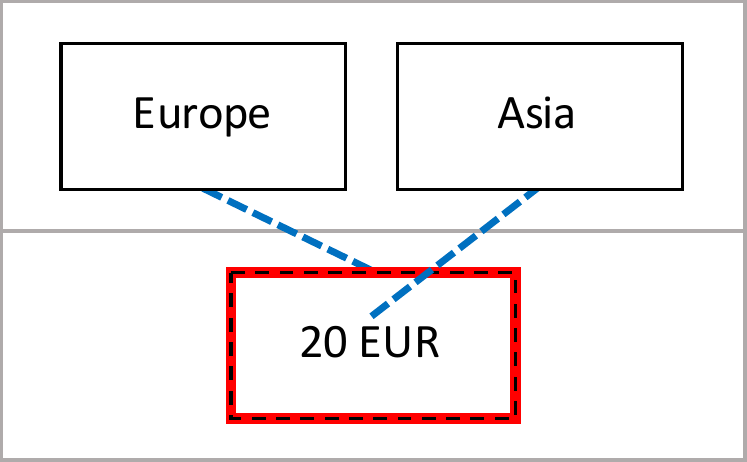}
      \label{sub:validation4}
    }
    \caption{The prototype features a built-in edge validation for error prevention.}
    \label{fig:validation}
  \end{center}
\end{figure}

Implementing the quality engineering
principle of \textit{poka-yoke}, the edge validation built into the user interface guarantees hypergraph validity by design~\cite{shingo1986zero}. When drawing the edges, visual feedback helps the user to avoid logical errors and enforces the connection rules as defined in Section~\ref{sec:contract_statement}. The different cases considered are displayed in Figure~\ref{fig:validation}.

For seamless integration in any conversational agent, the web application exposes an Application Programming Interface (API). As response format we chose the open-standard, light-weight JavaScript Object Notation (JSON). The parameter-value pairs are included as query string in the API url. 


\section{Conclusion}
\label{sec:conclusion}
We presented a top-down knowledge engineering approach which helps improve the capabilities of conversational agents. As opposed to the static answers of conventional FAQ bots, our approach enables smarter, dynamic responses. The user-friendly graphical user interface allows for rapid contract statement creation and updating with no technical skills required. The significant reduction of complexity cuts the cost to manage the contract statements, which represent the essential knowledge necessary to provide meaningful responses.

A further plus of the presented architecture is that the statement rules are exposed through a generic API. This renders the system framework-agnostic and makes it possible to integrate with any chatbot framework. Also, updates of the statement come into effect instantly. Finally, isolating the knowledge and the input validation from the chatbot framework keeps the latter light and lean.

In the future, we want to extend the keyword matching to more sophisticated methods: Fuzzy matching could render the matching process more robust towards typos. Making use of word embeddings could match synonyms which are not explicitly included in the keyword sets. Further, the data types of the keyword sets should be expanded to ranges of numbers and dates. Also, we see potential to refine the conversational flow by ranking the missing parameter values by probability and plausibility. Eventually, a language-agnostic architecture will be investigated.

\bibliography{references}
\bibliographystyle{IEEEtran}

\end{document}